\documentclass{bmvc2k}
\usepackage{amsmath}
\usepackage{bm}
\usepackage{multirow}
\usepackage{adjustbox}
\title{Unsupervised Clustering Active Learning for Person Re-identification}

\addauthor{Wenjing Gao}{wenjinggao@njust.edu.cn}{1}
\addauthor{Minxian Li$^\textbf{*}$}{minxianli@njust.edu.cn}{1}

\addinstitution{
School of Computer Science and Engineering\\
Nanjing University of Science and Technology
}

\runninghead{Student, Prof, Collaborator}{BMVC Author Guidelines}

\begin{document}

\maketitle

\begin{abstract}
Supervised person re-identification (re-id) approaches require a large amount of pairwise manual labeled data, which is not applicable in most real-world scenarios for re-id deployment.
On the other hand, unsupervised re-id methods rely on unlabeled data to train models but performs poorly compared with supervised re-id methods.
In this work, we aim to combine unsupervised re-id learning with a small number of human annotations to achieve a competitive performance. Towards this goal, we present a Unsupervised Clustering Active Learning (UCAL) re-id deep learning approach. It is capable of incrementally discovering the representative centroid-pairs and requiring human annotate them.
These few labeled representative pairwise data can improve the unsupervised representation learning model with other large amounts of unlabeled data.
More importantly, because the representative centroid-pairs are selected for annotation, UCAL can work with very low-cost human effort.
Extensive experiments demonstrate the superiority of the proposed model over state-of-the-art active learning methods on three re-id benchmark datasets.
\end{abstract}

\section{Introduction}
\label{sec:intro}
In recent years, person re-identification (re-id) attracted lots of research attentions because of its practical applications on public security and smart city \cite{gong2014person,li2014deepreid,zheng2015scalable,ristani2016performance}.
Person re-identification aims to recognize the same identity of person across non-overlapped cameras, which is a challenging task in computer vision.
Because of the non-overlap of ID labels between training and test set, re-id methods aims to learn a discriminative feature representation model for each person image.
%
Despite promising results reported in previous works, re-id relies heavily the acquisition of ID labels for each person image.
Unlike the labelling process for general categories which only requires each image to be labeled, \textit{the acquisition of ID labels need to annotate all pairs of person images, which costs huge human effort}.

Supervised re-id methods can achieve encouraging performances but require a large number of manually labeled identity matching image pairs.
However, the manual labeling for all person image pairs is a tedious and expensive process in re-id task. The cost of human labeling effort increases tremendously with the size of camera network and the number of persons in each camera.
One the other hand, unsupervised re-id methods can be trained by abundant unlabeled person images but significantly inferior in re-id accuracy.
Without cross-view pairwise ID labels, the re-id model is not able to learn the discriminative feature representation for the significant appearance change across cameras.

To save human labeling effort, active learning aims to select a subset of samples that are the most representative for labeling, and then use them to train the model in order to achieve a competitive performance compared with fully supervised models.
Specially, unsupervised active learning (UAL) \cite{li2020deep} aims to select representative samples from totally no labeled samples. It is also called early active learning (EAL) \cite{nie2013early}.
UAL/EAL re-id methods \cite{wang2016highly,liu2017early,roy2018exploiting,liu2019deep,liu2020pair} assume that no labeled image pairs are available in training set in the beginning.
Pairwise constraint minimization \cite{liu2017early}, triangle-free subgraph maximization \cite{roy2018exploiting}, and reinforcement learning \cite{liu2019deep} techniques are adopted to select a subset of pairwise samples for querying labeling.
After selecting a small number of representative data for labeling, existing unsupervised active learning use these pair annotation to train re-id model with reliable supervised models.
However, \textit{all the above active learning re-id methods train the re-id model only from the selected image pairs, losing sight of the rest unlabeled samples}.
Actually, the unlabeled data also benefit to re-id model to learn discriminative feature representation by exploring the association from the similar samples.

To overcome the above limitations, we consider a unsupervised clustering based active learning re-id framework.
Unsupervised clustering based deep model can offer a reliable data structure by generating pseudo labels without any human annotation.
Based on unsupervised clustering model, unsupervised active learning aims to select the representative sample pairs to reorganize the cluster structure.
The advantages are two-fold:
(1) Based on the clustering algorithm, unsupervised active learning method can easily and efficiently find the most representative sample pairs in the global space.
(2) Depending on the representative pairs selection, the reorganization of the cluster structure can help the feature representation learning in a more effective way.


In this work, the main contributions are as follows:
(1) We formulated an \textbf{U}nsupervised \textbf{C}lustering \textbf{A}ctive \textbf{L}earning (\textbf{UCAL}) model for person re-identification. This model combines jointly both unsupervised learning and active learning principles in an integrated learning framework. To the best of our knowledge, \textit{this is the first attempt at active learning with unsupervised learning person re-id model}.
(2) We propose an effective active learning strategy by the means of selecting the representative centroid-pairs from unsupervised clustering structure.
(3) We present a clustering reorganization method (i.e. splitting/merging) to maximize the effect of active learning, so it takes the low-cost human labeling labor.

Extensive comparative experiments demonstrate the advantages of UCAL
over the state-of-the-art active learning re-id approaches
on three popular benchmarks:
Market-1501\cite{zheng2015scalable}, DukeMTMC-ReID\cite{ristani2016performance, zheng2017unlabeled}, and MSMT17\cite{wei2018person}.
Especially, the proposed UCAL model achieves the best performance with the lowest annotation cost.


\section{Related Work}
\label{sec:related}

\noindent{\textbf{Unsupervised Learning in Re-ID.}}
According to whether auxiliary data is used in training stage, unsupervised learning methods can be devided into two groups:
(1) Pure unsupervsed learning. 
Most existing unsupervised learning re-id methods
\cite{chen2018deep,lin2019aBottom,lin2020unsupervised,wang2020unsupervised,zeng2020hierarchical} 
adopt clustering-based approaches to produce pseudo labels, and then update the feature representation model by these pseudo labels.
The challenge is how to obtain the precise cluster structure and how to alleviate the negative impact by noisy pseudo labels.
(2) Unsupervised domain adaptation (UDA). 
UDA approaches aim to transfer the learned knowledge from a labeled source domain to an unlabeled target domain.
The methods can be classified into three groups:
Source domain pre-trained methods \cite{fan2018unsupervised,zhang2019self,fu2019self,yang2020asymmetric,zhai2020ad}, 
Image-synthesis based methods \cite{wei2018person,deng2018image,chen2019instance,huang2019sbsgan,tang2020cgan}, 
and Joint-learning based methods \cite{peng2016unsupervised,zhong2018generalizing,zhong2019invariance,ge2020mutual,ge2020self}. 
Generally, the performance of unsupervised learning approaches is limited, because of the lack of the pair relation supervision.

\noindent{\textbf{Semi-Supervised Learning in Re-ID.}}
Semi-supervised learning methods \cite{liu2014semi,fu2019self,bak2017one,ye2018robust,wu2018exploit,li2018unsupervised,li2019unsupervised} 
train the re-id model both on pre-labeled data and unlabeled data.
Liu et al.\cite{liu2014semi} proposed a coupled dictionary learning by randomly pre-labeling one-third training data.
Most of these methods  \cite{bak2017one,ye2018robust,wu2018exploit,li2018unsupervised,li2019unsupervised}
work on one-shot learning setting. \textit{One-shot learning requires selecting all or most person identities, and then labeling one image or tracklet for each identity}.
Bak et al. \cite{bak2017one} learned a texture metric and a color metric on each camera pair by a one-shot metric learning approach.
Ye et al. \cite{ye2018robust} proposed a label estimation approach to learn feature representations by using the pre-labeled data to formulate an anchor graph.
Wu et al. \cite{wu2018exploit} initialized a CNN model using pre-labeled data per ID, and then adopted a step-wise learning approach to update the CNN model.
Li et al. \cite{li2018unsupervised, li2019unsupervised} used pre-labeled within-camera tracklet per ID to initialize a deep model, and then incrementally discover cross-camera tracklet association to improve the representation capability of deep model.

Although semi-supervised learning methods improve re-id performance by one-shot labeling strategy, this labeling setting is actually not practical for re-id task.
In practice, the total number of individual ID is hard to know, not mention to label one instance for each ID.

\noindent{\textbf{Active Learning and Human-in-the-loop in Re-ID.}}
For reducing the annotation cost in a more practical way, \emph{pair-wise labeling} re-id approaches are proposed.
According to the different applied stages, these methods can be divided into two categories, including active learning methods \cite{martinel2016temporal,wang2016highly,liu2017early,roy2018exploiting,liu2019deep,liu2020pair} and human-in-the-loop methods \cite{nie2013early,das2015active,wang2016human}:
(1) Active learning re-id methods aim at selecting a small number of image pairs to query human labeling in the training stage.
Liu et al. \cite{liu2017early} proposed an early active learning algorithm (EALPC) with a pairwise constraint to select the most representative samples for labeling. 
Roy et al. \cite{roy2018exploiting} presented a pairwise training subset selection framework to minimize human annotation effort.
Liu et al. \cite{liu2019deep} designed a deep reinforcement active learning (DRAL) model to minimize human effort in the training stage. 
(2) Human-in-the-loop re-id methods focus on optimising the ranking list of every probe by human feedback directly in the test stage.
Liu et al. \cite{nie2013early} learned a post-rank function for re-ordering the rank list during a re-identification process. 
Wang et al. \cite{wang2016human} proposed a distance metric learning method to incrementally optimise each new probe by human annotation in the deployment stage. 

Compared with unsupervised learning and semi-supervised re-id methods,active learning and human-in-the-loop re-id methods reasonably add a handful of pair-wise labels to improve the performance.
This paper follows active learning scheme, and our objective is to minimize human labeling effort to improve the performance by jointly unsupervised learning and active learning.


\section{Methodology}
\label{sec:method}

\begin{figure*}[t]
	\centering
	\includegraphics[width=\textwidth]{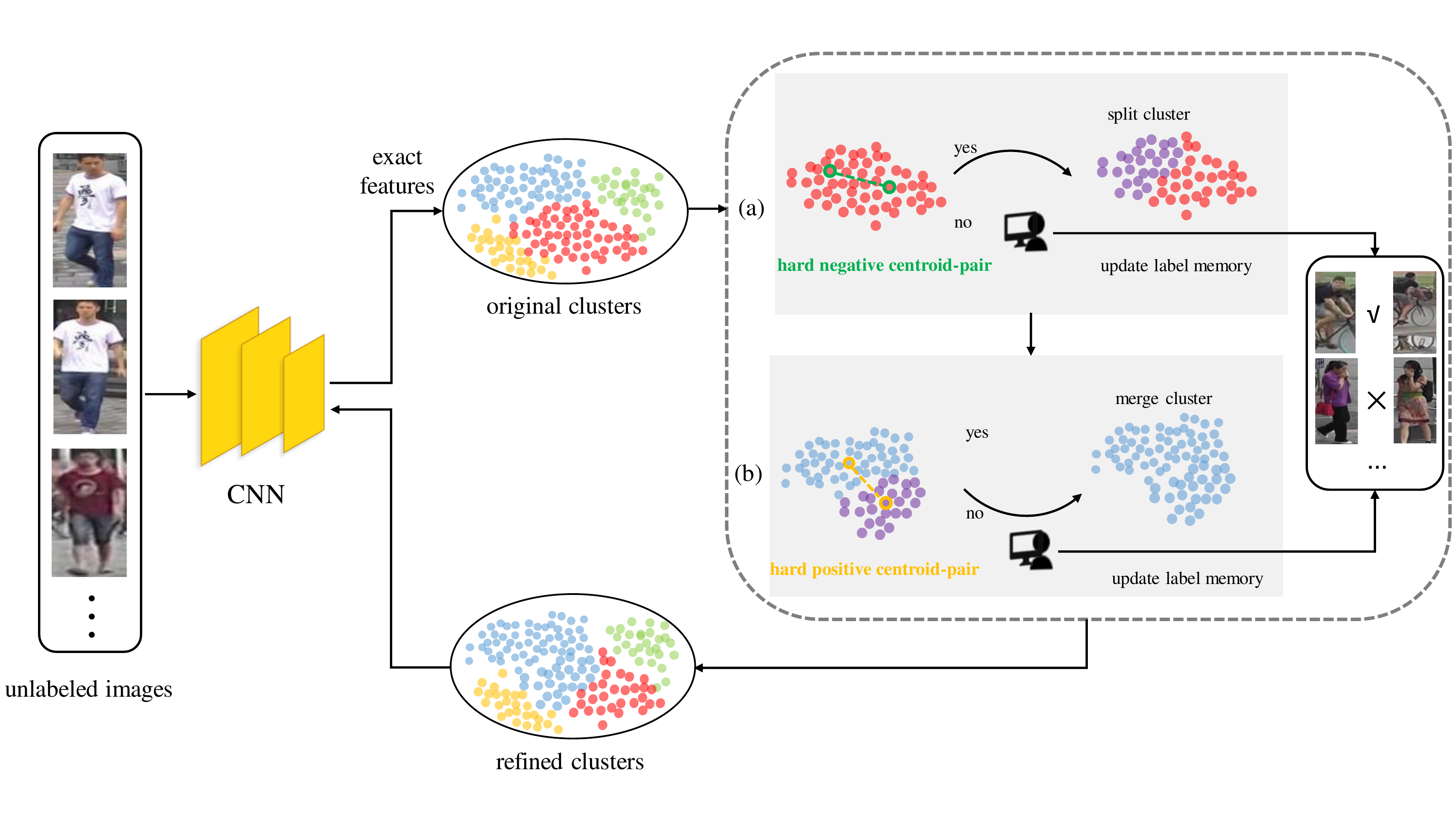}
	\caption{
		An overview of the proposed
		\textit{Unsupervised Clustering Active Learning} (UCAL) method for re-id model learning.
		The UCAL takes as unlabeled person images from all the camera views.
		The objective is to derive a person discriminative feature representative feature representation module by unsupervised active learning.
		To this end, we formulate the UCAL model (see Sec. \ref{subsec:ucal}) with
		(a) Split by Negative centroid-Pair Selection (SNPS) module
		and
		(b) Merge by Positive centroid-Pair Selection (MPPS) module in an unsupervised clustering framework.
		Different colours correspond to different clusters.
		Best viewed in colour.
	}
\label{fig:pipeline}
\end{figure*}

In this section, we introduce the overall framework of our Unsupervised Clustering Active Learning (UCAL) method for person re-identification.
An overview of the proposed UCAL model is depicted in Fig. \ref{fig:pipeline}.
Given $N$ unlabeled training person images $\mathcal{I}=\{\bm{I}_1,\bm{I}_2,\cdots,\bm{I}_N\}$ extracted from multiple camera views.
The training objective is to select $M$ pairs of unlabeled data to ask for human's labelling and learn a discriminative person re-id model.
The proposed UCAL framework is driven by both of unlablled data and labeled data.
In particular, we first adopt an unsupervised clustering method to discover the data distribution structure.
And then, we select the key \textit{centroid-pairs} from the clustering structure
by \textit{Negative centroid-Pair Selection (SNPS)} module and \textit{Merge by Positive centroid-Pair Selection (MPPS)} module,
and annotate the positive/negative relationships for these pairs.
At last, the model is updated by the re-organized clustering structure in an iterative way to learn a discriminative feature representation.

\subsection{Base Unsupervised Clustering Model}
\label{subsec:base}
We adopt a DBSCAN based unsupervised clustering method \cite{ester1996density,ge2020self} as the base clustering framework.
In this framework, a CNN-based \cite{he2016deep} encoder $f_{\theta}$ is used to represent person image feature.
Given the unlabeled training data $\mathcal{I}$, we adopt a self-paced clustering strategy \cite{ge2020self} to group the data into clusters.
According to the results of clustering, we update the parameters $\theta$ of the feature encoder $f_{\theta}$.
We enforce $||\bm{x}||=1$ via a L2-normalization layer.
The loss function is an unsupervised contrastive learning which is similar with \cite{ge2020self,wu2018unsupervised}, but computed only by the unlabeled data.
\begin{equation}
    \label{eq:loss}
    Loss(\bm{x}) = -log\frac{exp(\bm{x}^T\bm{c}_k/\tau)}
                            {\sum_{i=1}^{n}exp(\bm{x}^T\bm{c}_i/\tau)}
\end{equation}
where $\bm{x}^T\bm{c}_k$ expresses the similarity between the sample $\bm{x}$ and the $k$-th cluster centroid $\bm{c}_k$, $n$ is the number of clusters, and $\tau$ is a temperature parameter that controls the concentration of the distribution \cite{hinton2015distilling}.

\subsection{Unsupervised Active Learning Framework}
\label{subsec:ucal}
The base unsupervised clustering algorithm inevitably generates the incorrect pseudo labels, which will be harmful to the CNN model.
For refining the clustering results,
we aim to explore two kinds of representative sample pairs from the clustering structure:
\textbf{\textit{hard negative centroid-pair}}
and
\textbf{\textit{hard positive centroid-pair}}.
The \textit{hard negative centroid-pair} is the pair between two group centroids which have two IDs
but wrongly grouped into the same cluster.
The \textit{hard positive centroid-pair} is the pair between two cluster centroids which have the same ID
but wrongly grouped into two different clusters.
After obtaining these two kinds of sample pairs candidates, positive/negative relationships are required to label by human experts.
After that, two categories of manipulations:
\textbf{split} and \textbf{merge}
are adopted to refine the current clustering structure.
The CNN model is then updated by the refined clustering result.

\subsubsection{Split by Hard Negative Centroid-Pair Selection}
\label{ssub_snps}
A single cluster generated the base clustering algorithm may contain some small groups belonged to different ground truth IDs.
This kind of relationship between these groups is called \textit{hard negative}.
This result gives incorrectly the same pseudo label to these individual samples grouped into one cluster, which be harmful to update the feature representation CNN model.
To alleviate this problem, we propose a \textbf{S}plit by hard \textbf{N}egative centroid-\textbf{P}air \textbf{S}election (\textbf{SNPS}) method to mine the hard negative centroid-pairs within the same cluster.

Given $n_t$ clusters $\mathcal{C}=\{C_1,C_2,\cdots,C_{n_t}\}$ at epoch $t$,
we use $k$-medoids \cite{kaufman2009finding} algorithm to generate $k$ group candidates
for each cluster.
For $j$-th group $G_j$, $\bm{g}_j$
represents the centroid of $G_j$ (Fig.\ref{fig:pipeline}(a)).
%
However, the real number of groups which have negative relations depends on the different clusters.
That is, $k$ is not fixed for each cluster.
Therefore, we design a reliability criterion to decide $k$ for each cluster.
As discussed in \cite{ge2020self}, a reliable cluster should have two properties: high independence and high compactness.
Thus, we aim to find $k$ groups for each cluster with high independence and high compactness as splitting candidates.
According to this criterion, an algorithm is proposed to decide $k$ as follow:
\begin{equation}
    \label{eq:comp}
    comp_{j} = \min Sim_{intra}(G_j) \in [0,1]
\end{equation}
\begin{equation}
    \label{eq:comp}
    indep_{j} = 1-\max Sim_{inter}(G_j,G) \in [0,1]
\end{equation}
\begin{equation}
    \label{eq:k_select}
    k^* = \mathop{\arg\max}_{k\in[2,k_{max}]}
        {\sum_{j=1}^{k} comp_{j} \times indep_{j}}
\end{equation}
where
$Sim_{intra}$ represents the similarities between samples within group $G_j$,
$Sim_{inter}$ represents the similarities between samples from group $G_j$ and samples from other groups,
$k_{max}$ is set to $\frac{\sqrt{|C_i|/2}}{2}$.

\noindent{\textbf{Discussion.}}
Although the similar concept of independence and compactness is proposed in \cite{ge2020self}, the objective is quite different in this paper.
The objective of independence and compactness in \cite{ge2020self} is to make each cluster more reliable.
In SNPS, we aim to obtain a \textit{reliable splitting structure} by measuring the independence and compactness of several groups controlled by $k$.
Moreover, the criterion of good independence and compactness in \cite{ge2020self} is decided by the density threshold, which is difficult to choose appropriately. In this paper, we adopt the maximum product of independence and compactness to select the most reliable splitting structure.

\subsubsection{Merge by Hard Positive Centroid-Pair Selection}
\label{ssub_mpps}
The individual samples which have the same person ID are very likely separated into different clusters by the clustering algorithm.
This improper clustering structure generates the different pseudo labels to those clusters which should be grouped into the same cluster.
This kind of relationship between these clusters is called \textit{hard positive}.
The existence of hard positive pairs harms the representation ability of CNN model.
To mine the hard positive relationship between clusters, we propose a \textbf{M}erge by hard \textbf{P}ositive centroid-\textbf{P}iar \textbf{S}election (\textbf{MPPS}) method.

Given $n_t$ clusters $\mathcal{C}=\{C_1,C_2,\cdots,C_{n_t}\}$ at epoch $t$, the cluster centroid of cluster $C_i$ is defined as $\bm{c}_i$, and $s(\bm{c}_i,\bm{c}_j) \in [0,1]$ represents the similarity between $\bm{c}_i$ and $\bm{c}_j$.
For each centroid $\bm{c}_i$,
a rank list $[s_1,s_2,\cdots,s_l,\cdots,s_{l_{max}}]$
is computed by the similarity between $\bm{c}_i$ and other centroids \textit{from high to low}.
$l_{max}$ is the maximum number of candidate clusters.
If all centroid-pairs are provided to human experts, the cost will be very high.
To lower the labeling cost, we design a measure function to decide how many the similar centroid-pairs are selected to labeling.
We take the similarity difference between the adjacent centroids in the rank list as the measure value.
\begin{equation}
    \bm{d} = \{d_l=s_l-s_{l+1} | l \in [1,l_{max}-1]\}
\end{equation}
\begin{equation}
    \label{eq:l_select}
    d^*_l = \frac{d_l-min(\bm{d})}
               {max(\bm{d})-min(\bm{d})}
            \in [0,1]
\end{equation}
where $d^*_l$ is the min-max normalization value.

Inspired by \cite{rodriguez2014clustering}, the high value of $d_l^*$ indicates a large margin of similarity difference between $s_l$ and $s_{l+1}$.
It means the positive probabilities of pair $c_i$-$c_l$ and pair $c_i$-$c_{l+1}$ are uncertainty.
So, pair $c_i$-$c_l$ and pair $c_i$-$c_{l+1}$ need to be labeled by users.
We select the first $l$ centroids with $\bm{c}_j$ as the labeling pairs when $d^*_l > \delta$.
For the centroid-pairs labelled as positive, the corresponding clusters will be merged to one cluster, which is given the same pseudo label during the current training epoch.


\subsection{Overall Model Training}
We adopt a self-paced clustering strategy [10] as the base clustering algorithm, and the loss function (Eq. (\ref{eq:loss})) for the CNN parameters updating.
During the first 15 training epochs, the above pure unsupervised method works as the initialized model.
After the first 15 epochs of the training process, we deploy SNPS and MPPS modules to minimise the negative effect of unstable clustering structure.
Specifically, the UCAL model first deploy SNPS module (Sec. \ref{ssub_snps}) to find the hard negative centroid-pairs and split the original cluster into some smaller clusters.
And then, we deploy MPPS module (Sec. \ref{ssub_mpps}) to find the hard positive centroid-pairs and merge these clusters to a bigger one.
To lower the labeling cost, we also design a \textit{label memory} (Fig. \ref{fig:pipeline}) to record the positive/negative relationships between centroids labeled by human expert.
If the relationship between two centroids has been labeled in the label memory, the labeling request of this relationship is no need in the subsequent labeling process.



\section{Experiments}

\subsection{Datasets and Evaluation Protocol}
\label{sec:exp_dataset}

\noindent \textbf{Datasets.}
To evaluate the proposed UCAL model, we reports results on three large person re-identification datasets:
Market-1501\cite{zheng2015scalable}, DukeMTMC-ReID\cite{zheng2017unlabeled}, MSMT17\cite{wei2018person}.

(1) Market-1501\cite{zheng2015scalable}: Market-1501 is widely used large-scale re-id dataset.
It contains 32,668 images of 1,501 person identities from 6 camera views.
There are 12,936 images of 751 identities in training set,
and 3,368 queries of 750 identities are used as the query set to search the true match among the remained 19,732 images.

(2) DukeMTMC-ReID\cite{zheng2017unlabeled}: DukeMTMC-ReID is another one of the most popular large scale re-id dataset which consists 36,411 pedestrian images from 1,812 person identities captured from 8 different cameras.
Specifically, 16,522 images (702 identities) are adopted for training, 2,228 (702 identities) images are used as query to be retrieved from the remaining 17,661 images (1,110 identities).

(3) MSMT17\cite{wei2018person}: MSMT17 is a larger and more challenging dataset, which contains 4,101 identities and 126,441 pedestrian images.
There are 32,621 images of 1,041 identities in training set, and 93,820 images of 3,060 identities in test set. In the test set, 11,659 pedestrian images are randomly selected as probe images,
and the other 82,161 images are treated as gallery images.

\noindent \textbf{Evaluation Protocol.}
In the experiments, we used the Cumulative Matching Characteristic (CMC) and mean Average Precision (mAP) metrics to measure the methods' performance.
The cost of human labeling effort is computed as follow:
\begin{equation}
    \label{eq:cost}
    cost = \frac{M}{N*(N-1)/2}*100\%
\end{equation}
where $N$ represents the number of unlabeled training samples,
and $M$ represents the labeled pairs from human experts.

\subsection{Implementation Details}
\label{sec:exp_dataset}
We adopt an ImageNet pre-trained ResNet-50 \cite{he2016deep} as the backbone for our UCAL model.
After pooling-5 layer,we remove subsequent layers and add a
1D BatchNorm layer and an L2-normalization layer
to derive the feature representations.
Person bounding box images are resized to $256\times128$ as input.
To ensure each training mini-batch has person images from all cameras,
we set the batch size to 64 for all datasets mentioned in this paper.
We adopted Adam optimiser with a weight decay of 0.0005,
the learning rate is initialized to $3.5\times10^{-4}$.
We train the model for 50 epochs,
and the learning rate is divided by 10 after every 20 epochs.
From the 15th epoch and every subsequent epoch,we use SNPS and MPPS modules to generate labeled pairs.
By default, we set the maximum number of clusters that can be merged in each epoch to 20$\%$ of the total number of clusters and $\delta=0.3$.
%
All the experiments on three datasets follow the same settings as above.

\subsection{Comparisons with State-Of-The-Art Methods}
\label{sec:exp_sota}

In this section, we compare the proposed UCAL model with sixteen state-of-the-art re-id methods, including
unsupervised learning (OIM\cite{xiao2017joint},
BUC\cite{lin2019aBottom},
SSL\cite{lin2020unsupervised},
MMCL\cite{wang2020unsupervised},
HCT\cite{zeng2020hierarchical}),
unsupervised domain adaptation (PTGAN\cite{wei2018person},
ECN\cite{zhong2019invariance},
SSG\cite{fu2019self},
MMCL$_{trans}$\cite{wang2020unsupervised},
SPCL\cite{ge2020self}),
semi-supervised learning
(EUG\cite{wu2018exploit},
TAUDL\cite{li2018unsupervised},
UTAL\cite{li2019unsupervised},
SSG++\cite{fu2019self}),
and active learning
(TMA\cite{martinel2016temporal},
DRAL\cite{liu2019deep}) models.
%
The rank-(1,5,10) matching accuracy($\%$) and mAP($\%$) performance evaluated on
Market-1501 \cite{zheng2015scalable}, DukeMTMC-ReID \cite{zheng2017unlabeled}, MSMT17 \cite{wei2018person}
are showed in Table \ref{tab:SOTA}.
Moreover, the cost of human labeling effort computed by Eq.(\ref{eq:cost}) is also given.
The baseline model is trained by the unsupervised clustering method (Sec. \ref{subsec:base}) without any labeling cost.
The supervised model adopt the same loss function (Eq. (\ref{eq:loss})) for training but using ground truth label instead of pseudo label.
The experimental results show three observations as follows.

\begin{table*}[!t]
	\centering
	\caption{Comparison of proposed UCAL approach with state-of-the-art unsupervised, domain adaptation, semi-supervised and active learning approaches on Market-1501, DukeMTMC-ReID, and MSMT17.
	}
	\label{tab:SOTA}
	\begin{adjustbox}{width=\columnwidth,center}
	\begin{tabular}
		{|c||c|c||c|c|c|c|c||c|c|c|c|c||c|c|c|c|c|}
		\hline
		& \multicolumn{2}{c||}{\multirow{2}{*}{Methods}}
		& \multicolumn{5}{c||}{Market-1501 \cite{zheng2015scalable}}
		& \multicolumn{5}{c||}{DukeMTMC-ReID \cite{zheng2017unlabeled}}
		& \multicolumn{5}{c|} {MSMT17 \cite{wei2018person}} \\
		\cline{4-18}
		& \multicolumn{2}{c||}{} & R1 & R5 & R10 & mAP & cost(\%) 	& R1 & R5 & R10 & mAP & cost(\%) & R1 & R5 & R10 & mAP & cost(\%) \\
		\hline \hline
		\multirow{5}{*}{unsup}
		& OIM\cite{xiao2017joint}         & CVPR'17     & 38.0 & 58.0 & 66.3 & 14.0 & 0     & 24.5 & 38.8 & 46.0 & 11.3 & 0     & - & -	& -	& -	& - \\
		& BUC\cite{lin2019aBottom}        & AAAI'19     & 66.2 & 79.6 & 84.5 & 38.3 & 0     & 47.4 & 62.6 & 68.4 & 27.5 & 0     & - & -	& -	& -	& - \\
		& SSL\cite{lin2020unsupervised}   & CVPR'20     & 71.7 & 83.8 & 87.4 & 37.8 & 0     & 52.5 & 63.5 & 68.9 & 28.6 & 0     & - & -	& -	& -	& - \\
		& MMCL\cite{wang2020unsupervised} & CVPR'20     & 80.3 & 89.4 & 92.3 & 45.5 & 0     & 65.2 & 75.9 & 80.0 & 40.2 & 0     & 35.4 & 44.8 & 49.8 & 11.2 & 0 \\
		& HCT\cite{zeng2020hierarchical}  & CVPR'20     & 80.0 & 91.6 & 95.2 & 56.4 & 0     & 69.6 & 83.4 & 87.4 & 50.7 & 0     & - & -	& -	& -	& - \\
		\hline
		\multirow{5}{*}{domain}
		& PTGAN\cite{wei2018person}                 & CVPR'18      & 38.6 & - & 66.1 & - & 0           & 27.4 & - & 50.7 & - & 0           & 10.2 & - & 24.4 & 2.9 & 0 \\
		& ECN\cite{zhong2019invariance}             & CVPR'19      & 75.1 & 87.6 & 91.6 & 43.0 & 0     & 63.3 & 75.8 & 80.4 & 40.4 & 0     & 25.3 & 36.3 & 42.1 & 8.5 & 0 \\
		& SSG\cite{fu2019self}                      & ICCV'19      & 80.0 & 90.0 & 92.4 & 58.3 & 0     & 73.0 & 80.6 & 83.2 & 53.4 & 0     & 32.2 & - & 51.2 & 13.3 & 0  \\
		& MMCL$_{trans}$\cite{wang2020unsupervised} & CVPR'20      & 84.4 & 92.8 & 95.0 & 60.4 & 0     & 72.4 & 82.9 & 85.0 & 51.4 & 0     & 43.6 & 54.3 & 58.9 & 16.2 & 0 \\
		& SPCL\cite{ge2020self}                     & NIPS'20      & 89.7 & 96.1 & 97.6 & 77.5 & 0     & - & -	& -	& -	& -                & 53.7 & 65.0 & 69.8 & 26.8 & 0 \\
		\hline
		\multirow{4}{*}{semi}
		& EUG\cite{wu2018exploit} & CVPR'18 & 49.8 & 66.4 & 72.7 & 22.5 & -     & 45.2 & 59.2 & 63.4 & 24.5     & - & - & -	& -	& -	& -\\
		& TAUDL\cite{li2018unsupervised} & ECCV'18      & 63.7 & - & - & 41.2 & -           & 61.7 & - & - & 43.5 & -           & 28.4 & -	& -	& 12.5	& - \\
		& UTAL\cite{li2019unsupervised}  & PAMI'19      & 69.2 & - & - & 46.2 & -           & 62.3 & - & - & 44.6 & -           & 31.4 & -	& -	& 13.1	& - \\
		& SSG++\cite{fu2019self}         & ICCV'19      & 86.2 & 94.6 & 96.5 & 68.7 & -     & 76.0 & 85.8 & 89.3 & 60.3 & -     & 41.6  & - & 62.2 & 18.3 & -  \\
		\hline
		\multirow{3}{*}{active}
		& TMA\cite{martinel2016temporal} & ECCV'16   & 47.9  & - & - & 22.3 & 13.58        & - & - & - & - & -                             & - & -	& -	& -	& - \\
		& DRAL\cite{liu2019deep} & ICCV'19
		& \textcolor{blue}{\bf{84.2}}	
		& \textcolor{blue}{\bf{94.3}}	
		& \textcolor{blue}{\bf{96.6}}	
		& \textcolor{blue}{\bf{66.3}}
		& \textcolor{blue}{\bf{0.15}}	
		& \textcolor{blue}{\bf{74.3}}	
		& \textcolor{blue}{\bf{84.8}}	
		& \textcolor{blue}{\bf{88.4}}	
		& \textcolor{blue}{\bf{56.0}}	
		& \textcolor{blue}{\bf{0.12}}	
		& -	& -	& -	& - & - \\
		& DRAL$_{upper}$\cite{liu2019deep} & ICCV'19	
		& 88.0	& 95.3	& 96.8	& 73.3 & 100	& 78.0	& 88.7	& 91.6	& 60.9	& 100	        & -	& -	& -	& - & - \\
		\hline
		\multirow{3}{*}{ours}
		& Baseline      & this paper   & 87.1 & 95.0 & 96.5 & 69.9	& 0                     & 77.4 & 87.1 & 90.7 & 60.5	& 0             & 44.9 & 57.9 & 63.1 & 20.3	& 0 \\
		& Supervised   & this paper   & 94.2 & 98.2 & 98.9	& 83.4	& 100                       & 84.6 & 92.2	& 94.1	& 71.1	& 100               & 71.3 & 84.2	& 88.0	& 45.5	& 100 \\
		& \textbf{UCAL} & this paper   &\textcolor{red}{\bf{91.8}} &\textcolor{red}{\bf{96.8}} &\textcolor{red}{\bf{97.9}} &\textcolor{red}{\bf{78.2}} &\textcolor{red}{\bf{0.08}}	&\textcolor{red}{\bf{81.2}} &\textcolor{red}{\bf{89.7}} &\textcolor{red}{\bf{92.5}} &\textcolor{red}{\bf{66.3}} &\textcolor{red}{\bf{0.06}}		&\textcolor{red}{\bf{63.2}} &\textcolor{red}{\bf{75.1}} &\textcolor{red}{\bf{79.4}} &\textcolor{red}{\bf{35.7}} &\textcolor{red}{\bf{0.03}}	\\
		\hline
	\end{tabular}
	\end{adjustbox}
\vspace{-0.6cm}
\end{table*}

(1) The proposed UCAL model outperforms all competitors of active learning models with significant margins both on performance and cost.
For example,
the mAP margin is
11.9\%(78.2-66.3) on Market1501 and 10.3\%(66.3-56.0) on DukeMTMC-ReID.
More importantly, the labeling cost of UCAL model is only 0.08\% on Market1501
while TMA\cite{martinel2016temporal}'s cost is 13.58\%
and DRAL\cite{liu2019deep}'s cost is 0.15\%,
and 0.06\% on DukeMTMC-ReID
while DRAL\cite{liu2019deep}'s cost is 0.12\%.

(2) Compared with state-of-the-art unsupervised/domain adaptation/semi-supervised learning models, the performance of our model is superior but with very low cost.
%
Although there is no labeling cost under the above setting, the performance is limited especially on the large scale benchmark such as MSMT17.
As discussed in Section \ref{sec:related}, the labeling cost of semi-supervised learning is not able to estimated.
%
Compared with the state-of-the-art unsupervised learning model, our UCAL model improves the performance of rank1 and mAP on MSMT17 by 9.5\%(63.2-53.7) and 8.9\%(35.7-26.8) with 0.03\% cost.

(3) The proposed UCAL model narrows the gap between active learning and supervised learning models. Specifically, compared with the full labeling cost (100\%) requirement of supervised learning model, by our UCAL model, the gap of rank1/mAP is narrowed to
-2.4\%(91.8-94.2) and -5.2\%(78.2-83.4) with 0.08\% cost on Market1501,
-3.4\%(81.2-84.6) and -4.8\%(66.3-71.1) with 0.06\% cost on DukeMTMC-ReID,
-8.1\%(63.2-71.3) and -9.8\%(35.7-45.5) with 0.03\% cost on MSMT17.

\subsection{Component Analysis and Discussion}
\label{sec:exp_component}

We conducted detailed UCAL model component analysis on two large person re-id datasets, Market1501 and MSMT17.

\noindent \textbf{Effect of SNPS and MPPS Modules.}
We started by testing the performance impact of SNPS module and MPPS module.
Based on the baseline model (Section \ref{subsec:base}), we firstly tested our SNPS component and MPPS component separately. Then, we combined these two components and tested the final performance.
Table \ref{tab:component} shows that, the proposed SNPS and MPPS are both superior over the baseline model.
%
After combining these two modules together to refine the clustering result, the UCAL model achieves mAP gain of 7.2 percent (78.2-71.0) and 13.6 percent (34.8-21.2) on Market1501 and MSMT17 respectively.
This validates the proposed idea of split and merge modules, which refines the clusering structure by mining hard negative centroid-pairs and hard positive centroid-pairs.

\begin{table} [!h]
	\centering
	\caption{Effect of SNPS component and MPPS component.}
	\begin{adjustbox}{width=0.9\columnwidth,center}
	\label{tab:component}
	\begin{tabular}
		{|c||c|c|c|c|c||c|c|c|c|c|}
		\hline
		\multirow{2}{*}{Methods}
		&\multicolumn{5}{c||}{Market-1501 \cite{zheng2015scalable}}
		&\multicolumn{5}{c|}{MSMT17 \cite{wei2018person}} \\
		\cline{2-11}
		&R1 &R5 &R10 &mAP &cost(\%)  	&R1 &R5 &R10 &mAP &cost(\%) \\
		\hline \hline
		Baseline    &87.1 &94.6 &96.4 &71.0 &0    		&46.3 &60.0 &65.6 &21.2 &0     \\
		SNPS	    &90.5 &96.4 &97.7 &76.8 &0.027		&53.4 &66.1 &71.0 &26.7 &0.010 \\
		MPPS	    &88.9 &96.1 &97.6 &72.2 &0.046		&56.0 &68.7 &73.9 &30.2 &0.019 \\
		SNPS+MPPS &\bf{91.8} &\bf{96.8} &\bf{97.9} &\bf{78.2}	&0.075	    &\bf{63.2} &\bf{75.1} &\bf{79.4} &\bf{35.7} &0.033 \\
		\hline
	\end{tabular}
	\end{adjustbox}
\end{table}

\begin{figure}[!h]
\vspace{-0.5cm}
	\centering
	\subfigure[Market1501]{\includegraphics[width=0.48\linewidth]{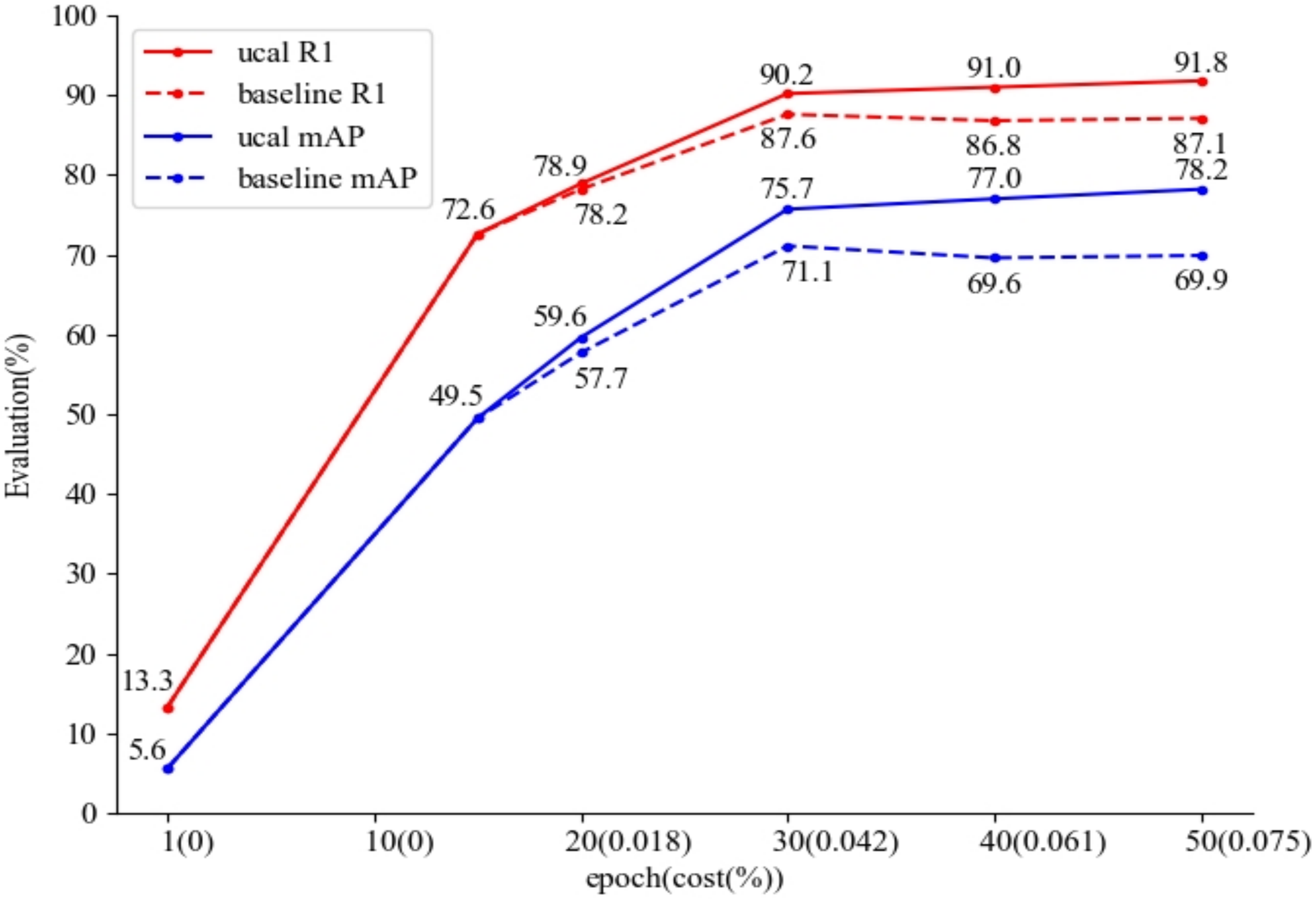}}
	\subfigure[MSMT17]{\includegraphics[width=0.48\linewidth]{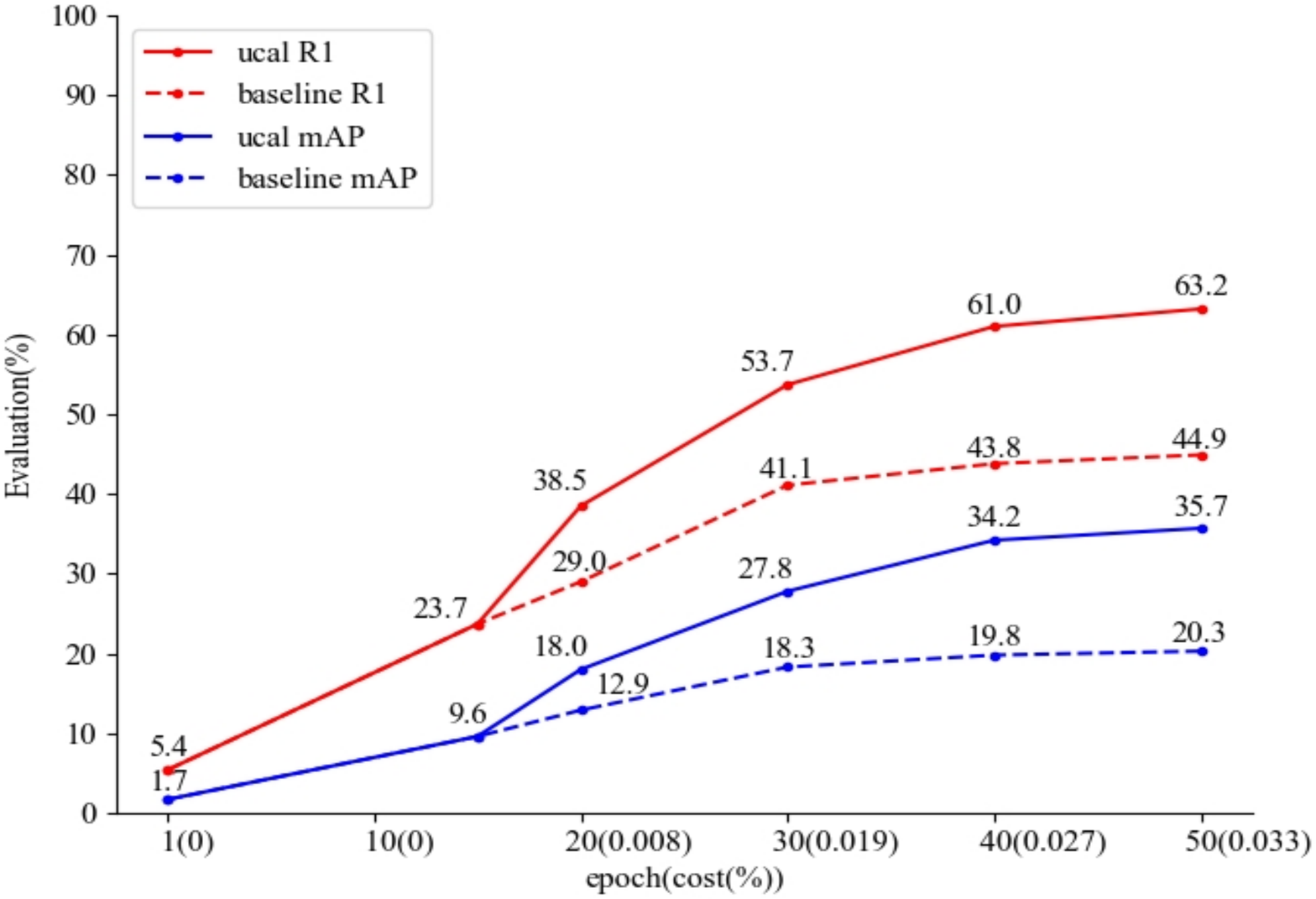}}
	\caption{The improvement of ucal compared with the base clustering model during training on (a) Market1501 and (b) MSMT17 benchmarks.
	The labeling cost of UCAL is depicted in the in brackets. Best viewed in colour.
	}
\label{fig:cost_effect}
\vspace{-0.5cm}
\end{figure}

\noindent \textbf{Effect with Labeling Cost.}
To further examine how well the proposed labeling strategy enables more discriminative re-id model learning, we tracked the improvement of UCAL compared with the base clustering model throughout the training.
Fig. \ref{fig:cost_effect} shows that, along with the increase of the labeling cost, the improvement of UCAL becomes more and more obvious versus the base clustering model.

\section{Conclusion}
We presented a novel \textit{Unsupervised Clustering Active Learning} (UCAL) model for active learning based person re-identification.
The model improves the feature representation ability of deep model dramatically with low labeling cost.
This is achieved optimising jointly both
Split by hard Negative centroid-Pair Selection (SNPS) module
and Merge by hard Positive centroid-Pair Selection (MPPS) module
by in a unified architecture.
Extensive evaluations were conducted on three large-scale person re-id benchmarks to validate the advantages of the proposed UCAL model over state-of-the-art active learning re-id methods.


\bibliography{bmvc}

\begin{thebibliography}{48}
\providecommand{\natexlab}[1]{#1}
\providecommand{\url}[1]{\texttt{#1}}
\expandafter\ifx\csname urlstyle\endcsname\relax
  \providecommand{\doi}[1]{doi: #1}\else
  \providecommand{\doi}{doi: \begingroup \urlstyle{rm}\Url}\fi

\bibitem[Bak and Carr(2017)]{bak2017one}
Slawomir Bak and Peter Carr.
\newblock One-shot metric learning for person re-identification.
\newblock In \emph{Proceedings of the IEEE conference on computer vision and
  pattern recognition}, pages 2990--2999, 2017.

\bibitem[Chen et~al.(2018)Chen, Zhu, and Gong]{chen2018deep}
Yanbei Chen, Xiatian Zhu, and Shaogang Gong.
\newblock Deep association learning for unsupervised video person
  re-identification.
\newblock \emph{Proc. Bri. Mach. Vis. Conf.}, 2018.

\bibitem[Chen et~al.(2019)Chen, Zhu, and Gong]{chen2019instance}
Yanbei Chen, Xiatian Zhu, and Shaogang Gong.
\newblock Instance-guided context rendering for cross-domain person
  re-identification.
\newblock In \emph{Proc. IEEE Int. Conf. Comput. Vis.}, pages 232--242, 2019.

\bibitem[Das et~al.(2015)Das, Panda, and Roy-Chowdhury]{das2015active}
Abir Das, Rameswar Panda, and Amit Roy-Chowdhury.
\newblock Active image pair selection for continuous person re-identification.
\newblock In \emph{IEEE Int. Conf. on Img. Proc.}, pages 4263--4267, 2015.

\bibitem[Deng et~al.(2018)Deng, Zheng, Ye, Kang, Yang, and Jiao]{deng2018image}
Weijian Deng, Liang Zheng, Qixiang Ye, Guoliang Kang, Yi~Yang, and Jianbin
  Jiao.
\newblock Image-image domain adaptation with preserved self-similarity and
  domain-dissimilarity for person re-identification.
\newblock In \emph{Proc. IEEE Conf. Comput. Vis. Pattern Recognit.}, pages
  994--1003, 2018.

\bibitem[Ester et~al.(1996)Ester, Kriegel, Sander, Xu,
  et~al.]{ester1996density}
Martin Ester, Hans-Peter Kriegel, J{\"o}rg Sander, Xiaowei Xu, et~al.
\newblock A density-based algorithm for discovering clusters in large spatial
  databases with noise.
\newblock In \emph{Kdd}, volume~96, pages 226--231, 1996.

\bibitem[Fan et~al.(2018)Fan, Zheng, Yan, and Yang]{fan2018unsupervised}
Hehe Fan, Liang Zheng, Chenggang Yan, and Yi~Yang.
\newblock Unsupervised person re-identification: Clustering and fine-tuning.
\newblock \emph{ACM Transactions on Multimedia Computing, Communications, and
  Applications}, 14\penalty0 (4):\penalty0 1--18, 2018.

\bibitem[Fu et~al.(2019)Fu, Wei, Wang, Zhou, Shi, and Huang]{fu2019self}
Yang Fu, Yunchao Wei, Guanshuo Wang, Yuqian Zhou, Honghui Shi, and Thomas~S
  Huang.
\newblock Self-similarity grouping: A simple unsupervised cross domain
  adaptation approach for person re-identification.
\newblock In \emph{Proc. IEEE Int. Conf. Comput. Vis.}, pages 6112--6121, 2019.

\bibitem[Ge et~al.(2020{\natexlab{a}})Ge, Chen, and Li]{ge2020mutual}
Yixiao Ge, Dapeng Chen, and Hongsheng Li.
\newblock Mutual mean-teaching: Pseudo label refinery for unsupervised domain
  adaptation on person re-identification.
\newblock \emph{Proc. Int. Conf. on Learn. Rep.}, 2020{\natexlab{a}}.

\bibitem[Ge et~al.(2020{\natexlab{b}})Ge, Chen, Zhu, Zhao, and Li]{ge2020self}
Yixiao Ge, Dapeng Chen, Feng Zhu, Rui Zhao, and Hongsheng Li.
\newblock Self-paced contrastive learning with hybrid memory for domain
  adaptive object re-id.
\newblock \emph{Proc. Neur. Info. Proc. Sys.}, 2020{\natexlab{b}}.

\bibitem[Gong et~al.(2014)Gong, Cristani, Yan, and Loy]{gong2014person}
Shaogang Gong, Marco Cristani, Shuicheng Yan, and Chen~Change Loy.
\newblock \emph{Person re-identification}.
\newblock Springer, 2014.

\bibitem[He et~al.(2016)He, Zhang, Ren, and Sun]{he2016deep}
Kaiming He, Xiangyu Zhang, Shaoqing Ren, and Jian Sun.
\newblock Deep residual learning for image recognition.
\newblock In \emph{Proc. IEEE Conf. Comput. Vis. Pattern Recognit.}, pages
  770--778, 2016.

\bibitem[Hinton et~al.(2015)Hinton, Vinyals, and Dean]{hinton2015distilling}
Geoffrey Hinton, Oriol Vinyals, and Jeff Dean.
\newblock Distilling the knowledge in a neural network.
\newblock \emph{arXiv preprint arXiv:1503.02531}, 2015.

\bibitem[Huang et~al.(2019)Huang, Wu, Xu, and Zhong]{huang2019sbsgan}
Yan Huang, Qiang Wu, JingSong Xu, and Yi~Zhong.
\newblock Sbsgan: Suppression of inter-domain background shift for person
  re-identification.
\newblock In \emph{Proc. IEEE Int. Conf. Comput. Vis.}, pages 9527--9536, 2019.

\bibitem[Kaufman and Rousseeuw(2009)]{kaufman2009finding}
Leonard Kaufman and Peter~J Rousseeuw.
\newblock \emph{Finding groups in data: an introduction to cluster analysis},
  volume 344.
\newblock John Wiley \& Sons, 2009.

\bibitem[Li et~al.(2020)Li, Ma, Kang, Yuan, Zhang, and Wang]{li2020deep}
Changsheng Li, Handong Ma, Zhao Kang, Ye~Yuan, Xiao-Yu Zhang, and Guoren Wang.
\newblock On deep unsupervised active learning.
\newblock \emph{Proc. Int. Jo. Conf. of Artif. Intell.}, 2020.

\bibitem[Li et~al.(2018)Li, Zhu, and Gong]{li2018unsupervised}
Minxian Li, Xiatian Zhu, and Shaogang Gong.
\newblock Unsupervised person re-identification by deep learning tracklet
  association.
\newblock In \emph{Proc. Eur. Conf. Comput. Vis.}, pages 737--753, 2018.

\bibitem[Li et~al.(2019)Li, Zhu, and Gong]{li2019unsupervised}
Minxian Li, Xiatian Zhu, and Shaogang Gong.
\newblock Unsupervised tracklet person re-identification.
\newblock \emph{IEEE Trans. Pattern Anal. Mach. Intell.}, 42\penalty0
  (7):\penalty0 1770--1782, 2019.

\bibitem[Li et~al.(2014)Li, Zhao, Xiao, and Wang]{li2014deepreid}
Wei Li, Rui Zhao, Tong Xiao, and Xiaogang Wang.
\newblock Deepreid: Deep filter pairing neural network for person
  re-identification.
\newblock In \emph{Proc. IEEE Conf. Comput. Vis. Pattern Recognit.}, pages
  152--159, 2014.

\bibitem[Lin et~al.(2019)Lin, Dong, Zheng, Yan, and Yang]{lin2019aBottom}
Yutian Lin, Xuanyi Dong, Liang Zheng, Yan Yan, and Yi~Yang.
\newblock A bottom-up clustering approach to unsupervised person
  re-identification.
\newblock In \emph{AAAI Conf. on Art. Intel.}, 2019.

\bibitem[Lin et~al.(2020)Lin, Xie, Wu, Yan, and Tian]{lin2020unsupervised}
Yutian Lin, Lingxi Xie, Yu~Wu, Chenggang Yan, and Qi~Tian.
\newblock Unsupervised person re-identification via softened similarity
  learning.
\newblock In \emph{Proc. IEEE Conf. Comput. Vis. Pattern Recognit.}, pages
  3390--3399, 2020.

\bibitem[Liu et~al.(2017)Liu, Chang, Chen, and Yang]{liu2017early}
Wenhe Liu, Xiaojun Chang, Ling Chen, and Yi~Yang.
\newblock Early active learning with pairwise constraint for person
  re-identification.
\newblock In \emph{Joint European Conference on Machine Learning and Knowledge
  Discovery in Databases}, pages 103--118, 2017.

\bibitem[Liu et~al.(2020)Liu, Chang, Chen, Phung, Zhang, Yang, and
  Hauptmann]{liu2020pair}
Wenhe Liu, Xiaojun Chang, Ling Chen, Dinh Phung, Xiaoqin Zhang, Yi~Yang, and
  Alexander~G Hauptmann.
\newblock Pair-based uncertainty and diversity promoting early active learning
  for person re-identification.
\newblock \emph{ACM Transactions on Intelligent Systems and Technology},
  11\penalty0 (2):\penalty0 1--15, 2020.

\bibitem[Liu et~al.(2014)Liu, Song, Tao, Zhou, Chen, and Bu]{liu2014semi}
Xiao Liu, Mingli Song, Dacheng Tao, Xingchen Zhou, Chun Chen, and Jiajun Bu.
\newblock Semi-supervised coupled dictionary learning for person
  re-identification.
\newblock In \emph{Proc. IEEE Conf. Comput. Vis. Pattern Recognit.}, pages
  3550--3557, 2014.

\bibitem[Liu et~al.(2019)Liu, Wang, Gong, Lu, and Tao]{liu2019deep}
Zimo Liu, Jingya Wang, Shaogang Gong, Huchuan Lu, and Dacheng Tao.
\newblock Deep reinforcement active learning for human-in-the-loop person
  re-identification.
\newblock In \emph{Proc. IEEE Int. Conf. Comput. Vis.}, pages 6122--6131, 2019.

\bibitem[Martinel et~al.(2016)Martinel, Das, Micheloni, and
  Roy-Chowdhury]{martinel2016temporal}
Niki Martinel, Abir Das, Christian Micheloni, and Amit~K Roy-Chowdhury.
\newblock Temporal model adaptation for person re-identification.
\newblock In \emph{Proc. Eur. Conf. Comput. Vis.}, pages 858--877, 2016.

\bibitem[Nie et~al.(2013)Nie, Wang, Huang, and Ding]{nie2013early}
Feiping Nie, Hua Wang, Heng Huang, and Chris Ding.
\newblock Early active learning via robust representation and structured
  sparsity.
\newblock In \emph{Proc. Int. Jo. Conf. of Artif. Intell.}, pages 1572--1578,
  2013.

\bibitem[Peng et~al.(2016)Peng, Xiang, Wang, Pontil, Gong, Huang, and
  Tian]{peng2016unsupervised}
Peixi Peng, Tao Xiang, Yaowei Wang, Massimiliano Pontil, Shaogang Gong, Tiejun
  Huang, and Yonghong Tian.
\newblock Unsupervised cross-dataset transfer learning for person
  re-identification.
\newblock In \emph{Proc. IEEE Conf. Comput. Vis. Pattern Recognit.}, pages
  1306--1315, 2016.

\bibitem[Ristani et~al.(2016)Ristani, Solera, Zou, Cucchiara, and
  Tomasi]{ristani2016performance}
Ergys Ristani, Francesco Solera, Roger Zou, Rita Cucchiara, and Carlo Tomasi.
\newblock Performance measures and a data set for multi-target, multi-camera
  tracking.
\newblock In \emph{Workshop of Eur. Conf. Comput. Vis.}, pages 17--35, 2016.

\bibitem[Rodriguez and Laio(2014)]{rodriguez2014clustering}
Alex Rodriguez and Alessandro Laio.
\newblock Clustering by fast search and find of density peaks.
\newblock \emph{science}, 344\penalty0 (6191):\penalty0 1492--1496, 2014.

\bibitem[Roy et~al.(2018)Roy, Paul, Young, and
  Roy-Chowdhury]{roy2018exploiting}
Sourya Roy, Sujoy Paul, Neal~E Young, and Amit~K Roy-Chowdhury.
\newblock Exploiting transitivity for learning person re-identification models
  on a budget.
\newblock In \emph{Proc. IEEE Conf. Comput. Vis. Pattern Recognit.}, pages
  7064--7072, 2018.

\bibitem[Tang et~al.(2020)Tang, Xi, Wang, Song, and Gao]{tang2020cgan}
Yingzhi Tang, Yang Xi, Nannan Wang, Bin Song, and Xinbo Gao.
\newblock Cgan-tm: A novel domain-to-domain transferring method for person
  re-identification.
\newblock \emph{IEEE Trans. Img. Proc.}, 29:\penalty0 5641--5651, 2020.

\bibitem[Wang and Zhang(2020)]{wang2020unsupervised}
Dongkai Wang and Shiliang Zhang.
\newblock Unsupervised person re-identification via multi-label classification.
\newblock In \emph{Proc. IEEE Conf. Comput. Vis. Pattern Recognit.}, pages
  10981--10990, 2020.

\bibitem[Wang et~al.(2016{\natexlab{a}})Wang, Gong, and Xiang]{wang2016highly}
Hanxiao Wang, Shaogang Gong, and Tao Xiang.
\newblock Highly efficient regression for scalable person re-identification.
\newblock In \emph{Proc. Bri. Mach. Vis. Conf.}, pages 1--8,
  2016{\natexlab{a}}.

\bibitem[Wang et~al.(2016{\natexlab{b}})Wang, Gong, Zhu, and
  Xiang]{wang2016human}
Hanxiao Wang, Shaogang Gong, Xiatian Zhu, and Tao Xiang.
\newblock Human-in-the-loop person re-identification.
\newblock In \emph{Proc. Eur. Conf. Comput. Vis.}, pages 405--422,
  2016{\natexlab{b}}.

\bibitem[Wei et~al.(2018)Wei, Zhang, Gao, and Tian]{wei2018person}
Longhui Wei, Shiliang Zhang, Wen Gao, and Qi~Tian.
\newblock Person transfer gan to bridge domain gap for person
  re-identification.
\newblock In \emph{Proc. IEEE Conf. Comput. Vis. Pattern Recognit.}, pages
  79--88, 2018.

\bibitem[Wu et~al.(2018{\natexlab{a}})Wu, Lin, Dong, Yan, Ouyang, and
  Yang]{wu2018exploit}
Yu~Wu, Yutian Lin, Xuanyi Dong, Yan Yan, Wanli Ouyang, and Yi~Yang.
\newblock Exploit the unknown gradually: One-shot video-based person
  re-identification by stepwise learning.
\newblock In \emph{Proc. IEEE Conf. Comput. Vis. Pattern Recognit.}, pages
  5177--5186, 2018{\natexlab{a}}.

\bibitem[Wu et~al.(2018{\natexlab{b}})Wu, Xiong, Stella, and
  Lin]{wu2018unsupervised}
Zhirong Wu, Yuanjun Xiong, X~Yu Stella, and Dahua Lin.
\newblock Unsupervised feature learning via non-parametric instance
  discrimination.
\newblock In \emph{Proc. IEEE Conf. Comput. Vis. Pattern Recognit.}, pages
  3733--3742, 2018{\natexlab{b}}.

\bibitem[Xiao et~al.(2017)Xiao, Li, Wang, Lin, and Wang]{xiao2017joint}
Tong Xiao, Shuang Li, Bochao Wang, Liang Lin, and Xiaogang Wang.
\newblock Joint detection and identification feature learning for person
  search.
\newblock In \emph{Proc. IEEE Conf. Comput. Vis. Pattern Recognit.}, pages
  3376--3385. IEEE, 2017.

\bibitem[Yang et~al.(2020)Yang, Li, Zhong, Luo, Sun, Cheng, Guo, Huang, Ji, and
  Li]{yang2020asymmetric}
Fengxiang Yang, Ke~Li, Zhun Zhong, Zhiming Luo, Xing Sun, Hao Cheng, Xiaowei
  Guo, Feiyue Huang, Rongrong Ji, and Shaozi Li.
\newblock Asymmetric co-teaching for unsupervised cross-domain person
  re-identification.
\newblock In \emph{AAAI Conf. on Art. Intel.}, volume~34, pages 12597--12604,
  2020.

\bibitem[Ye et~al.(2018)Ye, Lan, and Yuen]{ye2018robust}
Mang Ye, Xiangyuan Lan, and Pong~C Yuen.
\newblock Robust anchor embedding for unsupervised video person
  re-identification in the wild.
\newblock In \emph{Proc. Eur. Conf. Comput. Vis.}, pages 170--186, 2018.

\bibitem[Zeng et~al.(2020)Zeng, Ning, Wang, and Guo]{zeng2020hierarchical}
Kaiwei Zeng, Munan Ning, Yaohua Wang, and Yang Guo.
\newblock Hierarchical clustering with hard-batch triplet loss for person
  re-identification.
\newblock In \emph{Proc. IEEE Conf. Comput. Vis. Pattern Recognit.}, pages
  13657--13665, 2020.

\bibitem[Zhai et~al.(2020)Zhai, Lu, Ye, Shan, Chen, Ji, and Tian]{zhai2020ad}
Yunpeng Zhai, Shijian Lu, Qixiang Ye, Xuebo Shan, Jie Chen, Rongrong Ji, and
  Yonghong Tian.
\newblock Ad-cluster: Augmented discriminative clustering for domain adaptive
  person re-identification.
\newblock In \emph{Proceedings of the IEEE/CVF Conference on Computer Vision
  and Pattern Recognition}, pages 9021--9030, 2020.

\bibitem[Zhang et~al.(2019)Zhang, Cao, Shen, and You]{zhang2019self}
Xinyu Zhang, Jiewei Cao, Chunhua Shen, and Mingyu You.
\newblock Self-training with progressive augmentation for unsupervised
  cross-domain person re-identification.
\newblock In \emph{Proc. IEEE Int. Conf. Comput. Vis.}, pages 8222--8231, 2019.

\bibitem[Zheng et~al.(2015)Zheng, Shen, Tian, Wang, Wang, and
  Tian]{zheng2015scalable}
Liang Zheng, Liyue Shen, Lu~Tian, Shengjin Wang, Jingdong Wang, and Qi~Tian.
\newblock Scalable person re-identification: A benchmark.
\newblock In \emph{Proc. IEEE Conf. Comput. Vis. Pattern Recognit.}, pages
  1116--1124, 2015.

\bibitem[Zheng et~al.(2017)Zheng, Zheng, and Yang]{zheng2017unlabeled}
Zhedong Zheng, Liang Zheng, and Yi~Yang.
\newblock Unlabeled samples generated by gan improve the person
  re-identification baseline in vitro.
\newblock In \emph{Proc. IEEE Int. Conf. Comput. Vis.}, pages 3754--3762, 2017.

\bibitem[Zhong et~al.(2018)Zhong, Zheng, Li, and Yang]{zhong2018generalizing}
Zhun Zhong, Liang Zheng, Shaozi Li, and Yi~Yang.
\newblock Generalizing a person retrieval model hetero-and homogeneously.
\newblock In \emph{Proc. Eur. Conf. Comput. Vis.}, pages 172--188, 2018.

\bibitem[Zhong et~al.(2019)Zhong, Zheng, Luo, Li, and
  Yang]{zhong2019invariance}
Zhun Zhong, Liang Zheng, Zhiming Luo, Shaozi Li, and Yi~Yang.
\newblock Invariance matters: Exemplar memory for domain adaptive person
  re-identification.
\newblock In \emph{Proc. IEEE Conf. Comput. Vis. Pattern Recognit.}, pages
  598--607, 2019.

\end{thebibliography}
\end{document}